\documentclass{article} 
\usepackage[preprint]{colm2026_conference}

\usepackage{microtype}
\usepackage{hyperref}
\usepackage{url}
\usepackage{booktabs}
\usepackage{amsmath}
\usepackage{amssymb}
\usepackage{algorithm}
\usepackage{algorithmic}
\usepackage{multirow}
\usepackage{graphicx}
\usepackage{lineno}
\usepackage{tikz}
\usetikzlibrary{arrows.meta,positioning,calc,decorations.pathreplacing}

\title{PACE: Anytime-Valid Acceptance Tests for Self-Evolving Agents}

\author{ZayxShawn\\Independent Researcher}

\begin{document}

\ifcolmsubmission
\linenumbers
\fi

\maketitle
\lhead{}

\begin{abstract}
Self-evolving agents improve by repeatedly proposing modifications to their own prompts, skills, or workflows and keeping those that score higher on a small held-out set. Nearly all effort has gone into the \emph{proposer}; we identify the \emph{acceptor} (the rule that decides whether to commit a change) as the loop's silent weak point. Applied hundreds of times against the same noisy dev estimate, the ubiquitous ``keep it if the score went up'' rule is uncontrolled adaptive multiple testing. The agent effectively \emph{p-hacks itself}, accumulating \emph{false commits} that make it churn and drift rather than improve. We recast committing as a sequential hypothesis test and propose \textbf{PACE} (\emph{Paired Anytime-valid Commit Evaluation}), a training-free \emph{anytime-valid commit gate}: each candidate is compared to the incumbent on identical instances and committed only when a testing-by-betting e-process accumulates decisive evidence, stopping early to save evaluations and controlling \emph{each candidate's} false-commit probability at a user-set level even under optional stopping (a per-decision guarantee, not a run-level one). We evaluate this on Qwen2.5 agents (0.5B--3B) self-evolving at the prompt level on three tasks (GSM8K, SVAMP, ARC-Challenge): a deliberately minimal testbed that isolates the accept decision, and on which PACE cleanly separates real gains from noise. With a known beneficial edit hidden among noisy proposals, greedy acceptance commits $30$--$42\%$ false and $10$--$33\%$ harmful modifications, while PACE commits the real improvement and essentially \emph{nothing else} (0/5 audit-labelled false commits), matching greedy's held-out accuracy at sharply lower variance ($+0.74{\pm}0.04$ vs.\ $+0.54{\pm}0.30$ at 3B) and ${\sim}18\%$ lower evaluation cost. With a stochastic agent and \emph{no} real gain available, greedy commits $13$--$21$ spurious self-modifications per run ($72$--$100\%$ false), churning and degrading the most fragile agent by $4.9$ points, while PACE commits almost nothing and holds at baseline. Reliability of self-evolution depends on the acceptor, not only on the proposer.
\end{abstract}

\section{Introduction}

Self-evolving agents now rewrite their own prompts, induce reusable skills, edit their controlling code, or restructure multi-agent workflows, and they have made rapid progress \citep{adas2024,godelagent2024,dgm2025,voyager2023,dspy2024}. Yet nearly all of it comes from building better \emph{proposers}: mechanisms that generate candidate modifications. The decision that closes the loop, whether to \emph{commit} a proposed change, has been left to a single unexamined heuristic. This omission is where self-evolution most often goes wrong: a good acceptor keeps the genuine gains a proposer finds and rejects the rest, cheaply.

\begin{figure}[t]
\centering
\includegraphics[width=\linewidth]{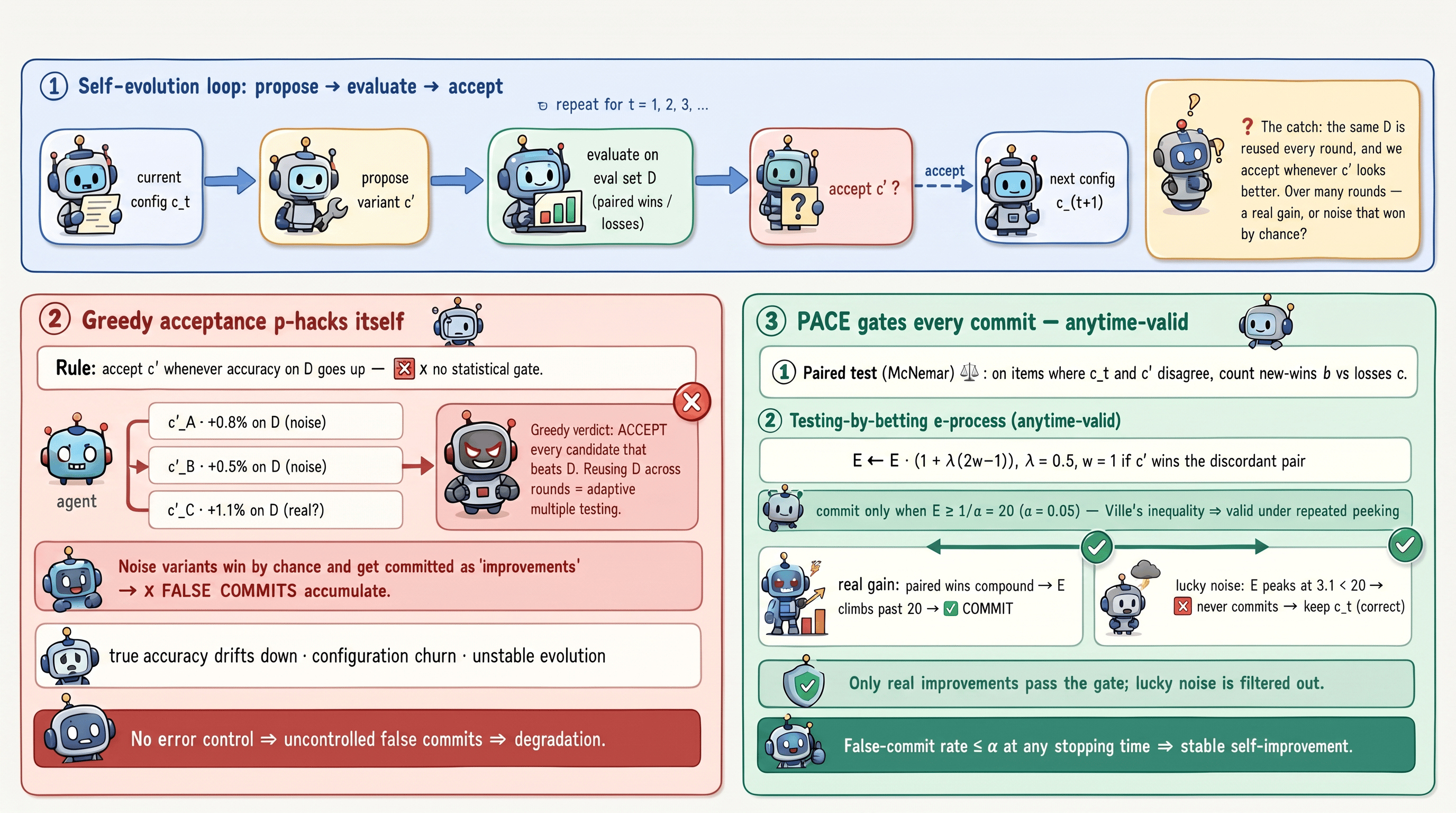}
\caption{\textbf{Overview of PACE.} A self-evolving agent proposes modifications and must decide whether to commit each. \textcolor{red!50!black}{Greedy} acceptance keeps any candidate whose reused dev score ticks up---\emph{p-hacking} a noisy, recycled signal. \textbf{PACE} instead runs an anytime-valid paired test (McNemar discordant pairs $+$ a betting e-process), committing only when evidence crosses $E\ge1/\alpha$, which bounds \emph{each candidate's} false-commit probability at $\alpha$ under optional stopping. A fresh held-out pool audits decisions for measurement only.}
\label{fig:overview}
\end{figure}

In practice the answer is overwhelmingly the same heuristic: measure the candidate on a small held-out set and keep it if the score went up. Some systems add lightweight safeguards (a second validation split, occasional human inspection, task-specific filters), but the operative rule that runs unattended, every round, remains ``commit iff the dev score improved.'' This paper makes a simple observation with sharp consequences. A self-evolution run applies this accept rule hundreds of times against the \emph{same}, \emph{noisy} estimate of quality. Statistically, this is adaptive multiple testing on a reused validation signal. Just as a researcher who tests many hypotheses on one dataset and keeps whatever reaches significance produces a flood of false discoveries, a self-evolving agent that keeps every change that bumped a small dev score accumulates \emph{false commits}: modifications that helped the estimate by chance but do not help (or actively hurt) true performance. The agent then churns. It constantly modifies itself, drifts, and at best wastes compute; at worst it degrades. A self-evolving agent with a greedy acceptor thus runs an unregistered, never-corrected sequence of trials against one tiny validation set and believes the winners.

We make this failure mode precise and give a simple statistical control for it. Our contributions:

\begin{itemize}
\item \textbf{A diagnosis.} We frame the commit step of self-evolution as a sequence of hypothesis tests and show that greedy acceptance is uncontrolled adaptive testing---predicting, and empirically exhibiting, a high rate of false and harmful commits (\S\ref{sec:problem}).
\item \textbf{A method (PACE).} We propose \emph{PACE}, an anytime-valid commit gate: a training-free wrapper that compares each candidate to the incumbent on identical instances and commits only when a testing-by-betting e-process clears a calibrated threshold, stopping early to minimize evaluations (\S\ref{sec:method}). The method is deliberately simple---a single paired sequential test, ${\sim}10$ lines---and applies to any self-modification loop whose incumbent and candidate can be scored on shared instances (binary correctness, or pairwise preferences), consuming only those paired outcomes rather than the proposer's internals. To our knowledge this is the first \emph{explicit} treatment of the accept step in agent self-evolution as an anytime-valid hypothesis test, building on a long line of sequential and safe-anytime-valid inference \citep{wald1947,testbybetting2021,gametheoreticstats2023}; the contribution is this \emph{abstraction}, locating an anytime-valid test at the commit step, not the test itself (which is standard).
\item \textbf{Evidence.} Across Qwen2.5 agents (0.5B--3B) self-evolving on three tasks spanning arithmetic (GSM8K, SVAMP) and multiple-choice science (ARC-Challenge), PACE drives greedy's $30$--$100\%$ audit-labelled false-commit rate to ${\approx}0\%$ and its $13$--$21$ noise commits per run to ${\approx}0$, while preserving genuine gains, in both a controlled regime with a known beneficial edit and a realistic stochastic regime---at lower evaluation cost than greedy (\S\ref{sec:exp}).
\end{itemize}

We study this acceptor in prompt-level self-evolution: deliberately the \emph{minimal} self-modification loop, where any churn or degradation traces to the accept rule rather than to a complex proposer or a brittle execution substrate. By interface the gate is proposer- and loop-agnostic, consuming only paired correctness outcomes, so richer loops (skills, code, multi-agent topologies) inherit the very same decision. We demonstrate it only on prompt evolution, however, and so treat system-level generality as argued, not shown.

\section{Related Work}

\paragraph{Self-evolving agents.} Recent systems search over agent designs, code, prompts, and skills, retaining candidates by benchmark performance: ADAS programs agents in code \citep{adas2024}, the G\"odel Agent rewrites its own logic \citep{godelagent2024}, the Darwin G\"odel Machine evolves self-modifying coding agents \citep{dgm2025}, Voyager grows a skill library \citep{voyager2023}, and Agent Workflow Memory induces reusable routines \citep{awm2024}; see \citet{selfevolvesurvey2025} for a survey. All select modifications by an empirical score and explicitly flag objective hacking and stability of self-modification as open problems; none treat acceptance as a statistical decision---the gap we fill.

\paragraph{Prompt and workflow optimization.} A parallel line optimizes the scaffold around a frozen model---few-shot demonstrations and instructions (DSPy \citep{dspy2024}), text ``gradients'' (TextGrad \citep{textgrad2024}), instruction search (OPRO \citep{opro2024}), reflective or evolutionary prompt search (GEPA \citep{gepa2025}, Promptbreeder \citep{promptbreeder2023}, EvoPrompt \citep{evoprompt2024}), and workflow search (AFlow \citep{aflow2025}). These optimize a scalar fitness on a fixed development set and are known to overfit it; we address the orthogonal question of whether a candidate should be committed at all.

\paragraph{Self-improvement and its pathologies.} Loop-based self-improvement (STaR \citep{star2022}, Reflexion \citep{reflexion2023}, Self-Refine \citep{selfrefine2023}, Self-Rewarding LMs \citep{selfreward2024}) is bounded by the reliability of self-evaluation: intrinsic self-correction can degrade reasoning \citep{cannotselfcorrect2024}, the generation--verification gap governs and eventually closes self-improvement \citep{mindthegap2025}, unanchored loops are reward-hacked and collapse \citep{canlrmselftrain2025}, and recursive self-training erodes diversity \citep{modelcollapse2024}. These analyze the \emph{signal} that self-improvement produces; we address the \emph{decision rule} that consumes it.

\paragraph{Sequential and anytime-valid testing.} PACE draws on safe, anytime-valid inference: e-processes and testing-by-betting yield nonnegative martingales whose maxima are controlled by Ville's inequality, enabling tests that remain valid under optional stopping \citep{wald1947,testbybetting2021,gametheoreticstats2023}; online error-control procedures bound error over streams of tests \citep{lordpp2017}. We import this machinery into the agent self-evolution loop, where, to our knowledge, commit decisions are currently made by unguarded point estimates.

\section{The Commit Decision in Self-Evolution}
\label{sec:problem}

\paragraph{Setup.} A self-evolving agent maintains a configuration $c_t$ (here, a system prompt) and runs for $T$ rounds. At round $t$ a proposer suggests a modification $c_t'$ (add/rewrite/delete/merge an instruction). An \emph{accept rule} $A$ decides whether to commit ($c_{t+1}=c_t'$) or reject ($c_{t+1}=c_t$). Quality is read through a small held-out set $D$ of size $n$ via accuracy $\hat{s}_D(c)$, an estimate of the agent's true accuracy $s(c)$.

\paragraph{Greedy acceptance is adaptive multiple testing.} The near-universal rule is greedy:
\begin{equation}
A_{\text{greedy}}: \;\text{commit} \iff \hat{s}_D(c_t') > \hat{s}_D(c_t).
\end{equation}
Because $D$ is small and reused every round, $\hat{s}_D$ is a noisy estimate of $s$, and over $T$ rounds the agent performs $T$ comparisons against the same noise. Consider what happens when no real improvement is available ($s(c_t')\le s(c_t)$ for every round): the candidate still wins the comparison with non-trivial probability simply because $\hat{s}_D$ fluctuates. Each such accepted change is a \emph{false commit}, and a \emph{harmful commit} if true accuracy strictly decreases. Greedy therefore accumulates false commits at a rate set by the dev-set noise, independent of whether real improvements exist---the agent keeps modifying itself on noise. This is precisely the structure of adaptive multiple testing on a reused dataset: the agent is running $T$ unregistered, uncorrected significance tests and keeping whatever happens to look good.

\paragraph{Why the obvious fixes fall short.} This is a textbook case of \emph{adaptive data analysis}: reusing one validation set to steer a long, data-dependent sequence of choices invalidates its naive estimates and inflates false discoveries \citep{reusableholdout2015,ladder2015}. The standard remedies sit poorly in an autonomous loop. A larger dev set only postpones the problem---its noise shrinks as $1/\sqrt{n}$ while the number of adaptive comparisons grows with the run. A fixed multiplicity correction (Bonferroni or $\alpha$-spending) must know the number of tests in advance, which an open-ended run does not, and spends its budget so fast that genuine gains are missed (\S\ref{sec:exp}). Refreshing the holdout every round restores validity but assumes a stream of fresh labeled data the agent rarely has. What the loop actually needs is a test that stays valid under an \emph{unbounded, adaptively chosen} number of looks at the same data---an anytime-valid sequential test, which is what the next section builds.

\paragraph{What we want.} A good accept rule should commit a modification only when there is reliable evidence that it improves true accuracy, while (i) keeping genuine improvements, (ii) spending as little evaluation as possible, and (iii) requiring no training. The next section gives such a rule.

\section{PACE: An Anytime-Valid Commit Gate}
\label{sec:method}

\paragraph{Paired evaluation.} A naive comparison of $\hat{s}_D(c_t)$ and $\hat{s}_D(c_t')$ conflates two sources of variance: which instances are easy, and whether the candidate is genuinely better. Pairing removes the first. We evaluate the incumbent $c_t$ and candidate $c_t'$ on the \emph{same} instances. For instance $i$ let $w_i=1$ if the candidate is correct and the incumbent is wrong, $w_i=0$ if the reverse, and discard ties (both right or both wrong)---a McNemar-style paired comparison \citep{mcnemar1947}. Under the null ``the candidate is not better,'' discordant pairs are equally likely either way: $\Pr[w_i=1]=\tfrac12$.

\paragraph{Testing by betting.} The idea behind an e-process is simple: treat testing as a betting game. We start with wealth $E_0=1$ and bet a fraction $\lambda$ of our wealth on each discordant pair being a ``win'' ($w_i=1$). If the null holds, the bet is fair and wealth stays near $1$ on average; if the candidate is truly better, wins outnumber losses and wealth grows. Concretely, after each discordant pair we update
\begin{equation}
E \leftarrow E\cdot\bigl(1+\lambda\,(2w_i-1)\bigr),\qquad \lambda\in[0,1).
\end{equation}
Under the null $\mathbb{E}[2w_i-1]=0$, so $E$ is a nonnegative martingale with $\mathbb{E}[E]=1$; by Ville's inequality $\Pr[\sup_t E_t\ge 1/\alpha]\le\alpha$ \citep{testbybetting2021,gametheoreticstats2023}. We therefore \textbf{commit} as soon as $E\ge 1/\alpha$, which controls the false-commit probability at level $\alpha$ \emph{at any stopping time}. Because the test is anytime-valid, we may evaluate instances incrementally and stop the moment the evidence is conclusive; if the evaluation budget is exhausted without crossing the threshold, the candidate is \textbf{rejected}. We use $\alpha=0.05$, $\lambda=0.5$, and a fixed batch size unless noted.

\paragraph{Guarantee.} The control is anytime-valid. Let $w_1,w_2,\dots$ be the discordant-pair outcomes in evaluation order and $\mathcal{F}_i=\sigma(w_1,\dots,w_i)$. Under the null $H_0$ that the candidate is not better---so $\Pr[w_i{=}1\mid\mathcal{F}_{i-1}]\le\tfrac12$---the wealth $E_i=\prod_{j\le i}\bigl(1+\lambda(2w_j{-}1)\bigr)$ is a nonnegative supermartingale with $E_0{=}1$ and $\mathbb{E}_{H_0}[E_i]\le1$. For the commit time $\tau=\inf\{i:E_i\ge1/\alpha\}$, Ville's inequality gives $\Pr_{H_0}[\tau<\infty]\le\alpha$: the probability of \emph{ever} committing a non-improving candidate is at most $\alpha$, for any $\alpha\in(0,1)$, any $\lambda\in[0,1)$, and any data-dependent stopping. Two scope points keep this honest. First, validity is \emph{per candidate}: for a fixed $c_t'$, the discordant outcomes form the evaluation stream and, under $H_0$, each is conditionally fair ($\Pr[w_i{=}1\mid\mathcal{F}_{i-1}]\le\tfrac12$), so $E_i$ is a supermartingale and \emph{that candidate's} false-commit probability is $\le\alpha$. This is not a run-level familywise bound: across many tested non-improving candidates the \emph{expected} number of false commits grows at most as $\alpha$ per candidate---but this already defuses greedy's pathology, whose per-candidate false rate is set by the dev-set noise (typically far above $\alpha$) and compounds every round. Second, the null concerns \emph{this} candidate against the incumbent on paired instances; the adaptive \emph{reuse} of $D$ to generate the \emph{next} candidate is greedy's problem, and because PACE gates each decision independently at $\alpha$ this per-decision control curbs it in practice. The theorem covers each candidate conditional on its own evaluation stream; we treat the loop-level adaptivity as an empirical matter, not a theoretical claim. What the guarantee does \emph{not} control is power: a genuine improvement is committed only if its evidence crosses $1/\alpha$ within the evaluation budget (\S\ref{sec:exp} and Table~\ref{tab:sens} probe this empirically).

\begin{algorithm}[t]
\caption{Self-evolution with the PACE commit gate}
\label{alg:gate}
\begin{algorithmic}[1]
\STATE \textbf{input:} agent config $c$, dev set $D$, level $\alpha$, bet $\lambda$
\FOR{$t=1$ \TO $T$}
  \STATE $c' \leftarrow \textsc{Propose}(c)$
  \STATE $E \leftarrow 1$
  \FOR{instance $i$ drawn from $D$}
     \STATE evaluate $c,c'$ on $i$; \textbf{if} tie \textbf{then continue}
     \STATE $w \leftarrow \mathbf{1}[c'\text{ right},\,c\text{ wrong}]$
     \STATE $E \leftarrow E\,(1+\lambda(2w-1))$
     \STATE \textbf{if} $E\ge 1/\alpha$ \textbf{then} $c\leftarrow c'$; \textbf{break} \COMMENT{commit}
  \ENDFOR
\ENDFOR
\STATE \textbf{return} $c$
\end{algorithmic}
\end{algorithm}

\paragraph{Why so simple, and why anytime-valid.} A long stream of tests, one might expect, should require an online false-discovery-rate correction across rounds. We found one unnecessary and, worse, over-conservative: because PACE gates each decision independently at $\alpha$ (\S\ref{sec:method}), reuse of $D$ does not compound the per-decision error the way it does for greedy, whereas a decaying-memory FDR variant was so cautious that it sometimes \emph{missed the genuine improvement entirely}. (As above, this is per-decision control, not a run-level familywise bound.) We therefore keep PACE as the single paired anytime-valid test of Algorithm~\ref{alg:gate}. Anytime-validity, not just pairing, is what earns its keep. A \emph{fixed}-$n$ paired test ties with PACE when the gain is obvious (Table~\ref{tab:controlled}), but it must fix $n$ in advance: too small misses marginal gains, too large wastes evaluation. The e-process instead adapts the number of pairs to the evidence, which is why PACE alone holds $0\%$ false at a near-constant $\Delta$ across the full $\alpha$/$n$ sweep (Table~\ref{tab:sens}) where the fixed-$n$ and FDR variants degrade. It adds no model calls beyond the evaluations greedy already performs, and typically fewer, because it stops early.

\section{Experiments}
\label{sec:exp}

Our experiments answer three questions. \textbf{(Q1)} Does greedy acceptance actually accumulate false and harmful commits, as the diagnosis predicts? \textbf{(Q2)} Does PACE prevent them while keeping genuine improvements, and at what evaluation cost? \textbf{(Q3)} Is the effect robust---across model scales, across tasks and domains, and across PACE's own knobs? A controlled regime and a sensitivity sweep answer Q1--Q2 and the knob part of Q3; a stochastic regime and two additional tasks answer the rest.

\paragraph{Agents and task.} The agent is a frozen Qwen2.5-Instruct model \citep{qwen25} (0.5B, 1.5B, 3B) steered by an editable system prompt; it solves GSM8K grade-school math \citep{gsm8k2021}, scored by exact match against the dataset's ground-truth answer. A proposer (a separate LLM) suggests one prompt edit per round. We use disjoint splits: a small reused \texttt{dev} ($n{=}40$) and a fresh \texttt{audit} pool ($n{=}120$) used \emph{only} for measurement, which the accept rule never sees. All numbers are averages over 5 seeds.

\paragraph{Metrics.} For every decision we audit the change on a large, fresh audit pool ($n{=}120$, disjoint from dev, which the accept rule never sees). A commit is a \emph{false commit} if its audited accuracy change is $\le 0$, and \emph{harmful} if $<0$. We report the false- and harmful-commit rates among commits, the end-to-end accuracy change $\Delta$ measured on the audit pool, and the evaluation cost (number of dev problems scored). Two caveats keep this honest. The audit is a finite estimate, so these are \emph{observed} (audit-labelled) false commits, not ground truth---but auditing at temperature $0$ on a pool far larger than dev makes the reported gaps (greedy $30$--$100\%$ vs.\ gates $0\%$) dwarf its sampling error. And the audit measures the \emph{realized} outcome, whereas $\alpha$ bounds the \emph{prospective} false-commit probability (\S\ref{sec:method}); the two coincide here.

\paragraph{Two regimes.} The two play complementary roles---a \emph{diagnostic} with a known ground-truth answer (a unit test for acceptors), and a \emph{realistic} stochastic deployment with no planted gain; neither is a full agent-\emph{system} deployment, which we leave to future work. (i) \textbf{Controlled}: we seed the agent with one deliberately harmful instruction that the proposer can remove for a large, \emph{known} true gain; the run thus contains exactly one big real improvement amid the proposer's genuinely noisy edits. This isolates whether an accept rule \emph{keeps the real gain while rejecting noise}. (ii) \textbf{Stochastic}: the agent is sampled at temperature $0.7$ (as deployed agents are), so the dev signal is genuinely noisy even though no large improvement is available; this is the realism evidence---it tests whether an accept rule \emph{avoids chasing sampling noise}. In the stochastic regime the audit is measured at temperature $0$ (true quality) while the agent's own dev signal is sampled.

\subsection{Controlled: keep the real gain, reject noise}

Table~\ref{tab:controlled} compares greedy, a fixed-$n$ paired test, and our anytime-valid gate (5 seeds; ${\pm}$ is SD across seeds). The contrast is direct: greedy commits $3$--$3.4$ changes per run, $30$--$42\%$ false and $10$--$33\%$ harmful; both statistical rules commit exactly the one real improvement and \emph{nothing else} ($0\%$ false, $0\%$ harmful). At 1.5B all three rules reach the same held-out gain ($\Delta{\approx}{+}0.57$), so here the gate's advantage is hygiene rather than accuracy---it keeps the real improvement while committing none of greedy's noise, at lower cost. At 3B the difference is one of \emph{variance}: the gate reaches a tight $\Delta{=}{+}0.74{\pm}0.04$ while greedy is a high-variance ${+}0.54{\pm}0.30$, because greedy's occasional harmful commits cause large drops on some seeds. We therefore read the gate's value as \emph{reliability}---it removes the downside---rather than a guaranteed mean margin (the $+0.20$ gap is within seed noise at $n{=}5$). The gate also costs the least: by stopping each comparison once the evidence is conclusive it uses ${\sim}18\%$ fewer dev evaluations than greedy (e.g.\ $1712$ vs.\ $2080$ paired problems at 1.5B), though greedy runs no test at all. The weakest model (0.5B) cannot do GSM8K even with the handicap removed, so no real gain exists for it in this regime; we use 0.5B only in the stochastic regime below.

To check that the gate is not merely exploiting a large, obvious gain, we repeated the 1.5B run with a \emph{milder} handicap (a soft discouragement of reasoning rather than a hard override), leaving a smaller true gain of ${\approx}{+}0.18$. The gate still captures most of it ($\Delta{=}{+}0.14$ at $0\%$ false), while greedy takes the slightly larger raw gain ($+0.18$) at the cost of $17\%$ false and $17\%$ harmful commits. This is the expected precision/recall trade-off: with a weaker signal the anytime-valid test gives up a little recall to preserve its zero-false-commit guarantee.

\begin{table}[t]
\centering
\setlength{\tabcolsep}{6pt}
\small
\begin{tabular}{llrrrr}
\toprule
Model & Accept rule & Cmt & False & Harm & $\Delta$ \\
\midrule
\multirow{3}{*}{1.5B}
 & Greedy            & $3.4_{\pm1.2}$ & 42\% & 33\% & $+0.57_{\pm.03}$ \\
 & Paired ($n$) & 1.0 & 0\%  & 0\%  & $+0.57_{\pm.04}$ \\
 & \textbf{PACE} & 1.0 & \textbf{0\%} & \textbf{0\%} & $+0.57_{\pm.04}$ \\
\midrule
\multirow{3}{*}{3B}
 & Greedy            & $3.0_{\pm2.6}$ & 30\% & 10\% & $+0.54_{\pm.30}$ \\
 & Paired ($n$) & 1.0 & 0\%  & 0\%  & $+0.74_{\pm.04}$ \\
 & \textbf{PACE} & 1.0 & \textbf{0\%} & \textbf{0\%} & $+0.74_{\pm.04}$ \\
\bottomrule
\end{tabular}
\caption{\textbf{Controlled regime} (5 seeds; ${\pm}$ is SD). Greedy commits many false/harmful changes; the gate keeps only the genuine improvement (0/5 audit-labelled false commits). At 1.5B all rules tie on held-out $\Delta$ (the gate wins on hygiene + cost); at 3B the gate matches greedy's mean with far lower variance ($\pm.04$ vs $\pm.30$). Cmt: commits per run. False/Harm are \emph{audit-estimated} (retrospective, temp-0); the gate's $\alpha$ bounds the \emph{prospective} false-commit probability.}
\label{tab:controlled}
\end{table}

\subsection{Stochastic: don't chase sampling noise}

\begin{figure}[t]
\centering
\includegraphics[width=0.78\linewidth]{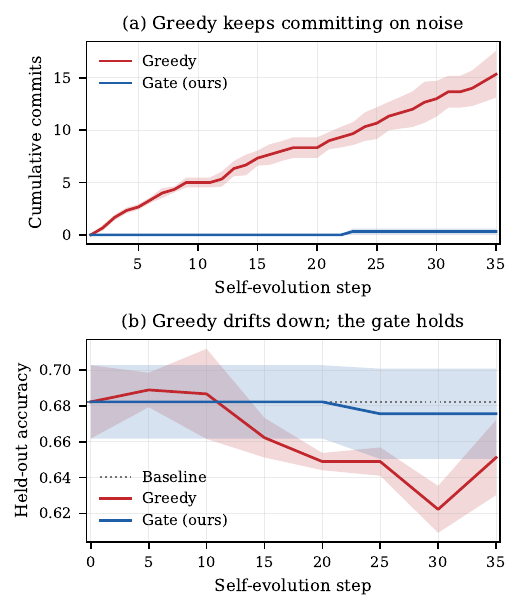}
\caption{\textbf{Stochastic-regime dynamics} (Qwen2.5-1.5B; mean$\pm$SE over 3 seeds). \textbf{(a)} Greedy reads sampling up-fluctuations as wins and commits continually ($\approx$15 edits over the run), while the gate commits essentially nothing. \textbf{(b)} Those commits are not free: greedy's true held-out accuracy drifts below where it started, whereas the gate holds at the baseline line.}
\label{fig:dynamics}
\end{figure}

Table~\ref{tab:stoch} reports the realistic regime and Figure~\ref{fig:dynamics} shows the dynamics. A sampled agent makes the dev estimate fluctuate, and greedy reads up-fluctuations as improvements: it commits a stream of $13.3$--$20.7$ self-modifications per run (SD $\le3.9$), $72$--$100\%$ of which a temperature-$0$ audit deems false. By contrast \emph{every} statistical gate---paired fixed-$n$, online-FDR, and ours---commits ${<}1$ per run: the family uniformly refuses to chase the noise, with greedy the lone offender. This commit-rate gap is the robust headline. The churn is not always harmless: on the most fragile agent (0.5B) the accumulated noise-chasing edits drive true held-out accuracy \emph{down} by $4.9{\pm}3.0$ points; at 1.5B the mean drop ($3.1$ points) is real but seed-noisy (${\pm}7.0$), and the robust 3B agent escapes net damage despite a $100\%$ false-commit rate. The \emph{guaranteed} cost of greedy, then, is the churn itself---continual, unjustified self-modification, a stability and compute liability in deployment---while held-out degradation is an additional risk that materializes on weaker agents. The gate holds held-out accuracy at baseline throughout (Figure~\ref{fig:dynamics}).

\begin{table}[t]
\centering
\small
\begin{tabular}{llrrr}
\toprule
Model & Accept rule & Commits & False & $\Delta$ \\
\midrule
\multirow{4}{*}{0.5B}
 & Greedy & $20.7_{\pm1.7}$ & 82\% & $-0.05_{\pm.03}$ \\
 & Paired ($n$) & 0.3 & 33\% & $+0.00$ \\
 & Online-FDR & 0.0 & 0\% & $+0.00$ \\
 & \textbf{PACE} & \textbf{0.3} & 33\% & $+0.00$ \\
\midrule
\multirow{4}{*}{1.5B}
 & Greedy & $15.3_{\pm3.9}$ & 72\% & $-0.03_{\pm.07}$ \\
 & Paired ($n$) & 0.7 & 67\% & $+0.00$ \\
 & Online-FDR & 0.0 & 0\% & $+0.00$ \\
 & \textbf{PACE} & \textbf{0.3} & 33\% & $-0.01$ \\
\midrule
\multirow{4}{*}{3B}
 & Greedy & $13.3_{\pm2.1}$ & 100\% & $\phantom{+}0.00$ \\
 & Paired ($n$) & 0.0 & 0\% & $\phantom{+}0.00$ \\
 & Online-FDR & 0.0 & 0\% & $\phantom{+}0.00$ \\
 & \textbf{PACE} & \textbf{0.0} & \textbf{0\%} & $\phantom{+}0.00$ \\
\bottomrule
\end{tabular}
\caption{\textbf{Stochastic regime}, agent sampled at $T{=}0.7$ (3 seeds; ${\pm}$ is SD). Greedy commits $13$--$21$ self-modifications/run ($72$--$100\%$ false by a temp-$0$ audit) and degrades the fragile 0.5B agent; every statistical gate commits ${<}1$ and holds $\Delta$ at baseline. False\% is \emph{audit-estimated} (retrospective), not the prospective $\alpha$ bound, and for the gates is over their $<\!1$ commit/run (a near-empty denominator).}
\label{tab:stoch}
\end{table}

\subsection{Sensitivity}

The gate exposes two knobs---the level $\alpha$ and the dev size $n$---and a natural worry is that its zero false-commit rate is merely bought by setting them conservatively. Table~\ref{tab:sens} (Appendix~\ref{app:extra}) shows otherwise. Sweeping $\alpha$ over a $10\times$ range ($0.01$ to $0.10$) and $n$ over a $4\times$ range ($20$ to $80$), the gate commits the single real improvement at \emph{every} setting, with $0\%$ false commits and a near-constant held-out gain $\Delta\approx{+}0.57$; greedy stays $33$--$53\%$ false throughout. The anytime-valid e-process is what buys this stability: because it accumulates evidence across instances rather than thresholding one fixed-$n$ estimate, it neither misses the gain when $\alpha$ is tightened nor fires on noise when $n$ is small. A fixed-schedule online-FDR alternative behaves far worse---it lost the genuine improvement entirely at $\alpha{=}0.01$ ($\Delta{=}{+}0.000$) and recovered only part of it at $n{=}20$ ($\Delta{=}{+}0.41$)---which is precisely why our gate is a single betting test rather than a multiplicity correction.

\subsection{Generalization across tasks and domains}

To test that the diagnosis and the fix are not artifacts of GSM8K, we repeat both regimes on SVAMP \citep{svamp2021} (a second arithmetic benchmark) and ARC-Challenge \citep{arc2018} (multiple-choice science---a \emph{non-math} domain). Table~\ref{tab:gen} (Appendix~\ref{app:extra}) shows the same pattern throughout. In the stochastic regime greedy commits $12$--$16$ self-modifications per run ($67$--$93\%$ false) and ends at or below baseline, while the gate commits essentially none and holds. In the controlled regime the gate again captures the planted improvement at $0\%$ false and matches greedy's held-out gain (e.g.\ $+0.45$ vs.\ $+0.45$ on SVAMP-3B, where greedy pays with $11\%$ false commits). The effect is thus a property of the \emph{accept rule under noisy evaluation}, not of any single task or domain.

\section{Discussion and Limitations}

The gate helps precisely where a self-evolution loop faces exploitable evaluation noise (a small or reused dev set, or a stochastic agent), which is the common case. When the base configuration is already near-optimal and the task is not prompt-sensitive, no rule (including greedy) commits much: on un-handicapped GSM8K, where instruct-tuned Qwen2.5 already sits near its prompt-induced ceiling, greedy and the gate alike commit essentially nothing ($\Delta{\approx}0$), so the gate's only effect is to cost nothing when there is nothing to decide. The danger the gate removes is the asymmetric one---churning on noise---and it removes it without sacrificing genuine gains.

\paragraph{When (not) to use it.} The price of the gate's safety is power, and that power scales with effect size: a large true gain crosses the e-process threshold in a handful of discordant pairs, a marginal gain needs many, and a gain too small for the evaluation budget to resolve is rejected---exactly the milder-handicap case, where the gate kept $78\%$ of a ${+}0.18$ improvement rather than all of it. The gate is therefore the wrong tool when genuine gains are expected to be tiny and a \emph{missed} improvement is costlier than a wrongly committed one; there, a higher $\alpha$ or a larger dev budget trades precision back for recall. In the far more common regime---noisy, reused evaluation where spurious commits are the dominant failure---it is a sensible default, at no greater cost than greedy. Finally, the gate controls false commits but inherits the dev set's coverage: it cannot certify improvements the dev set cannot see.

\begin{table}[t]
\centering\small
\setlength{\tabcolsep}{4pt}
\begin{tabular}{@{}p{0.28\linewidth}p{0.24\linewidth}p{0.38\linewidth}@{}}
\toprule
Claim & Support & Scope / caveat \\
\midrule
Per-candidate false-commit prob.\ $\le\alpha$ & Thm.\ (\S\ref{sec:method}) & Conditional on the candidate; not a run-level familywise bound \\
Greedy $30$--$100\%$ false; PACE ${\approx}0$ & Tables~\ref{tab:controlled}--\ref{tab:gen} & Audit-labelled (temp-$0$), not ground truth \\
Removes churn / degradation & Tab.~\ref{tab:stoch}, Fig.~\ref{fig:dynamics} & Strongest on small agents; $3$--$5$ seeds \\
Proposer-/loop-agnostic & Interface (paired outcomes) & Argued, not shown; prompt-level, one model family \\
\bottomrule
\end{tabular}
\caption{What PACE claims, what supports each claim, and its scope. The guarantee is per-decision; breadth across systems and model families is argued from the interface, not demonstrated.}
\label{tab:claims}
\end{table}

\paragraph{What PACE does and does not claim.} Table~\ref{tab:claims} summarizes the contract: PACE provides a per-decision false-commit bound under the paired null, empirically eliminates greedy's $30$--$100\%$ false commits and the resulting degradation, and claims neither higher power, a new statistic, nor system-level generality beyond its interface. Three caveats remain. We embed PACE in our own prompt-evolution loop rather than a third-party system (DSPy, ADAS); the interface is identical, but cross-system validation remains future work. Evidence uses one model family and one proposer, so proposer-agnosticism is argued at the interface level, not demonstrated. And greedy's held-out \emph{degradation} is strongest on small, fragile agents; the universally robust effect is false-commit suppression, not always a large accuracy rescue.

\section{Conclusion}

Self-evolving agents spend almost all of their effort proposing changes and almost none deciding whether to keep them; the unguarded commit step then accumulates false acceptances on a reused dev set. We made this failure mode precise (the commit step is uncontrolled adaptive testing) and showed that \emph{PACE}, a training-free paired anytime-valid test, bounds \emph{each candidate's} false-commit probability at a user-set level (per decision, not run-level) while preserving genuine gains, at lower cost than greedy. The acceptor is an overlooked piece of the self-evolution loop.

\bibliography{references}
\bibliographystyle{colm2026_conference}

\appendix
\section{Additional Results}
\label{app:extra}
Full sensitivity (Table~\ref{tab:sens}) and cross-task generalization (Table~\ref{tab:gen}) results, summarized in \S\ref{sec:exp}, are collected here.

\begin{table}[t]
\centering
\small
\begin{tabular}{lcccc}
\toprule
 & \multicolumn{2}{c}{Greedy} & \multicolumn{2}{c}{PACE} \\
\cmidrule(lr){2-3}\cmidrule(lr){4-5}
Setting & False & $\Delta$ & False & $\Delta$ \\
\midrule
$\alpha{=}0.01$ & 53\% & $+0.561$ & \textbf{0\%} & $\mathbf{+0.578}$ \\
default & 42\% & $+0.567$ & \textbf{0\%} & $\mathbf{+0.571}$ \\
$\alpha{=}0.10$ & 53\% & $+0.561$ & \textbf{0\%} & $\mathbf{+0.578}$ \\
$n{=}20$ & 33\% & $+0.574$ & \textbf{0\%} & $\mathbf{+0.578}$ \\
$n{=}80$ & 42\% & $+0.609$ & \textbf{0\%} & $\mathbf{+0.578}$ \\
\bottomrule
\end{tabular}
\caption{\textbf{Sensitivity} (1.5B, controlled regime, 3 seeds; default $\alpha{=}0.05$, $n{=}40$). Over a $10\times$ range of $\alpha$ and a $4\times$ range of dev size $n$, the gate commits the one real improvement with $0\%$ false commits and an essentially constant $\Delta$, while greedy remains $33$--$53\%$ false. The gate is insensitive to both knobs. (False\% is audit-estimated; $\alpha$ bounds the prospective rate.)}
\label{tab:sens}
\end{table}

\begin{table}[t]
\centering
\setlength{\tabcolsep}{6pt}
\small
\begin{tabular}{llcccc}
\toprule
 & & \multicolumn{2}{c}{Greedy} & \multicolumn{2}{c}{PACE} \\
\cmidrule(lr){3-4}\cmidrule(lr){5-6}
Task (regime) & Sz & Cmt/False & $\Delta$ & Cmt/False & $\Delta$ \\
\midrule
SVAMP (ctrl)  & 1.5B & 5.3 / 43\% & $+0.41$ & 1.0 / \textbf{0\%} & $+0.43$ \\
SVAMP (ctrl)  & 3B   & 2.7 / 11\% & $+0.45$ & 1.0 / \textbf{0\%} & $+0.45$ \\
SVAMP (stoch) & 1.5B & 12.0 / 74\% & $-0.01$ & \textbf{0.0} / \textbf{0\%} & $\phantom{+}0.00$ \\
ARC (stoch)   & 1.5B & 16.0 / 67\% & $-0.03$ & \textbf{0.0} / \textbf{0\%} & $+0.00$ \\
ARC (stoch)   & 3B   & 14.0 / 93\% & $-0.01$ & \textbf{0.0} / \textbf{0\%} & $\phantom{+}0.00$ \\
\bottomrule
\end{tabular}
\caption{\textbf{Generalization} to a second arithmetic task (SVAMP) and a non-math task (ARC-Challenge), 3 seeds. Controlled (ctrl): the gate keeps the planted gain at $0\%$ false. Stochastic (stoch): greedy churns $12$--$16$ edits/run ($67$--$93\%$ false) and drifts down, the gate commits ${\approx}0$ and holds. Cmt: commits/run; False\% is audit-estimated ($\alpha$ bounds the prospective rate).}
\label{tab:gen}
\end{table}

\end{document}